\documentclass[letterpaper]{article} % DO NOT CHANGE THIS
\usepackage{aaai2026}  % DO NOT CHANGE THIS
\usepackage{times}  % DO NOT CHANGE THIS
\usepackage{helvet}  % DO NOT CHANGE THIS
\usepackage{courier}  % DO NOT CHANGE THIS
\usepackage[hyphens]{url}  % DO NOT CHANGE THIS
\usepackage{graphicx} % DO NOT CHANGE THIS
\usepackage{natbib}  % DO NOT CHANGE THIS AND DO NOT ADD ANY OPTIONS TO IT
\usepackage{caption} % DO NOT CHANGE THIS AND DO NOT ADD ANY OPTIONS TO IT
\usepackage[linesnumbered,ruled,vlined]{algorithm2e}
\usepackage{booktabs}
\usepackage{multirow}
\usepackage{amsmath}%
\usepackage{xcolor}
\usepackage{orcidlink}

\title{A Generic Complete Anytime Beam Search for Optimal Decision Tree}
 \author {
     Harold Silvère Kiossou\orcidlink{0000-0001-6972-9885}\textsuperscript{\rm 1},
     Siegfried Nijssen\orcidlink{0000-0003-2678-1266}\textsuperscript{\rm 1,2},
     Pierre Schaus\orcidlink{0000-0002-3153-8941}\textsuperscript{\rm 1}
}
 \affiliations {
     \textsuperscript{\rm 1}UCLouvain, ICTEAM, Louvain-la-Neuve, Belgium\\
     \textsuperscript{\rm 2}KU LEUVEN, DTAI, Leuven, Belgium \\
     harold.kiossou@uclouvain.be , siegfried.nijssen@kuleuven.be, pierre.schaus@uclouvain.be
 }

\begin{document}

\maketitle

\begin{abstract}
Finding an optimal decision tree that minimizes classification error is known to be NP-hard. 
While exact algorithms based on MILP, CP, SAT, or dynamic programming guarantee optimality, they often suffer from poor anytime behavior—meaning they struggle to find high-quality decision trees quickly when the search is stopped before completion—due to unbalanced search space exploration. 
To address this, several anytime extensions of exact methods have been proposed, such as LDS-DL8.5, Top-$k$-DL8.5, and Blossom, but they have not been systematically compared, making it difficult to assess their relative effectiveness.
In this paper, we propose CA-DL8.5, a generic, complete, and anytime beam search algorithm that extends the DL8.5 framework and unifies some existing anytime strategies. 
In particular, CA-DL8.5 generalizes previous approaches LDS-DL8.5 and Top-$k$-DL8.5, by allowing the integration of various heuristics and relaxation mechanisms through a modular design. 
The algorithm reuses DL8.5’s efficient branch-and-bound pruning and trie-based caching, combined with a restart-based beam search that gradually relaxes pruning criteria to improve solution quality over time.
Our contributions are twofold:
(1) We introduce this new generic framework for exact and anytime decision tree learning, enabling the incorporation of diverse heuristics and search strategies;
(2) We conduct a rigorous empirical comparison of several instantiations of CA-DL8.5—based on Purity, Gain, Discrepancy, and Top-$k$ heuristics—using an anytime evaluation metric called the primal gap integral.
Experimental results on standard classification benchmarks show that CA-DL8.5 using LDS (limited discrepancy) consistently provides the best anytime performance, outperforming both other CA-DL8.5 variants and the Blossom algorithm while maintaining completeness and optimality guarantees.
\end{abstract}

\begin{links}
    \link{Code}{https://anonymous.4open.science/r/cadl85/}
    \link{Datasets}{https://dtai-static.cs.kuleuven.be/CP4IM/datasets/}
    % \link{Extended version}{https://aaai.org/example/extended-version}
\end{links}

\section{Introduction}
Decision trees are a fundamental machine learning model, widely adopted for their interpretability and solid performance in domains such as healthcare, finance, and education. Classic algorithms like CART~\cite{breiman1984cart} and C4.5~\cite{quinlan2014c45} induce decision trees greedily, selecting splits in a top-down manner based on local heuristics. These methods are fast but lack optimality guarantees, and they often produce suboptimal trees.

Recent years have seen growing interest in exact decision tree learning algorithms, which aim to find globally optimal trees, typically minimizing classification error or another loss function. These algorithms leverage combinatorial optimization techniques from MILP~\cite{bertsimas2017optimal, aghaei2021strong}, constraint programming~\cite{verhaeghe2020learning}, SAT~\cite{narodytska2018learning}, and dynamic programming~\cite{aglin2020learning, demirovic2022murtree}. While these methods offer strong generalization properties~\cite{bertsimas2017optimal, van2024optimal}, they tend to suffer from poor anytime behavior: when interrupted before convergence, they often return poor-quality solutions.

\begin{figure}[t]
\centering
\includegraphics[width=0.8\linewidth]{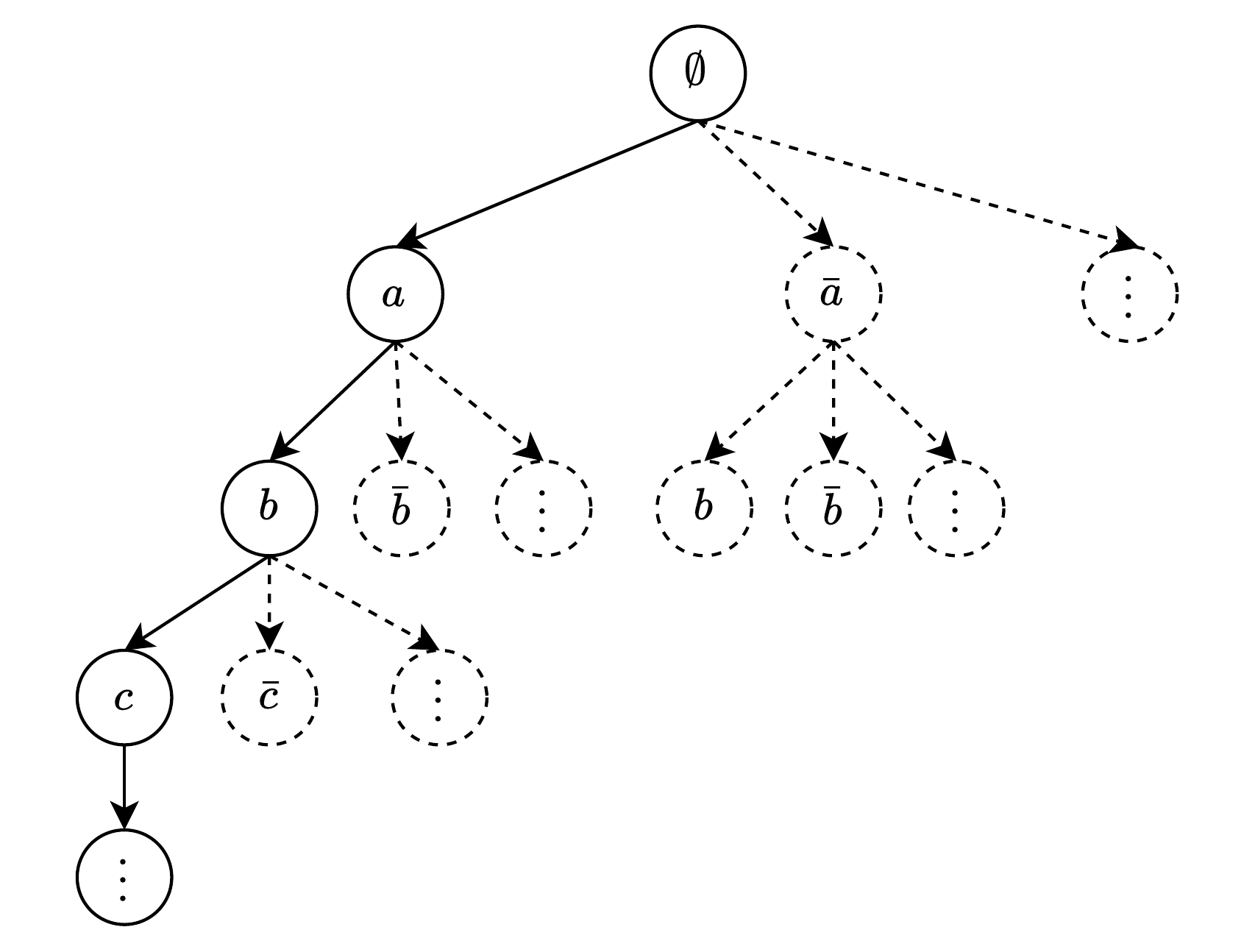}
\caption{DL8.5 explores the leftmost branches first, often leading to poor anytime performance when interrupted early.}
\label{fig:stuck_search}
\end{figure}

The DL8.5 algorithm~\cite{aglin2020learning}, based on dynamic programming and efficient caching, is a state-of-the-art method for optimal tree induction. DL8.5 explores the search space in a depth-first fashion, which often causes it to become stuck in unpromising regions, as illustrated in Figure~\ref{fig:stuck_search}. As a result, it may return poor trees when stopped early. Greedy methods like C4.5 provide quick results but lack the capacity to improve or guarantee optimality over time. Neither approach offers the benefits of a true anytime algorithm, which should return a good initial solution quickly and improve it continuously as time allows.

To improve the anytime performance or scalability of exact methods, three notable work have been proposed. LDS-DL8.5~\cite{kiossou2022time} integrates limited discrepancy search (LDS) into DL8.5, resulting in an algorithm that is both anytime and complete. Top-$k$-DL8.5~\cite{blanc2024harnessing} modifies DL8.5 by restricting the candidate features at each node to the Top-$k$ according to a ranking heuristic. This is a compromise between C4.5 and DL8.5: faster and more scalable, but unable to guarantee convergence to the optimal tree.
Finally, the Blossom algorithm~\cite{demirovic2023blossom} follows a fundamentally different search strategy. It uses a depth-first approach that expands decision tree nodes level by level, offering improved anytime behavior by avoiding the possible result of highly unbalanced trees when interrupted early as for DL8.5. Blossom is guaranteed to find the optimal tree given sufficient time.
Despite their promise, these three approaches have not been systematically compared in prior work, making it difficult to assess their relative strengths.

\paragraph{This paper makes two main contributions.}
First, we introduce CA-DL8.5, a novel algorithm for decision tree learning that is:
\textbf{complete} (guarantees optimality when given sufficient time),
\textbf{anytime} (produces high-quality solutions early and improves them over time),
and \textbf{generic} (easily instantiated with different heuristics and strategies).
CA-DL8.5 builds upon the Complete Anytime Beam Search (CABS) framework~\cite{zhang1998complete}, extending DL8.5 with an iterative weakening strategy that gradually relaxes pruning constraints across restarts. It generalizes LDS-DL8.5 and extends Top-$k$-DL8.5 to a complete method.

Second, we perform a rigorous empirical study of CA-DL8.5 under four heuristic strategies: Purity, Gain, Discrepancy, and Top-$k$. We evaluate these variants using the primal gap integral metric \cite{BERTHOLD2013611}, a principled anytime evaluation measure that captures performance over time. Our results show that CA-DL8.5 instantiated with discrepancy-based search (equivalent to LDS-DL8.5) and Top-$k$ consistently outperforms other instantiations and also improves upon the Blossom algorithm.

\begin{figure*}[ht]
\centering
\includegraphics[width=\linewidth]{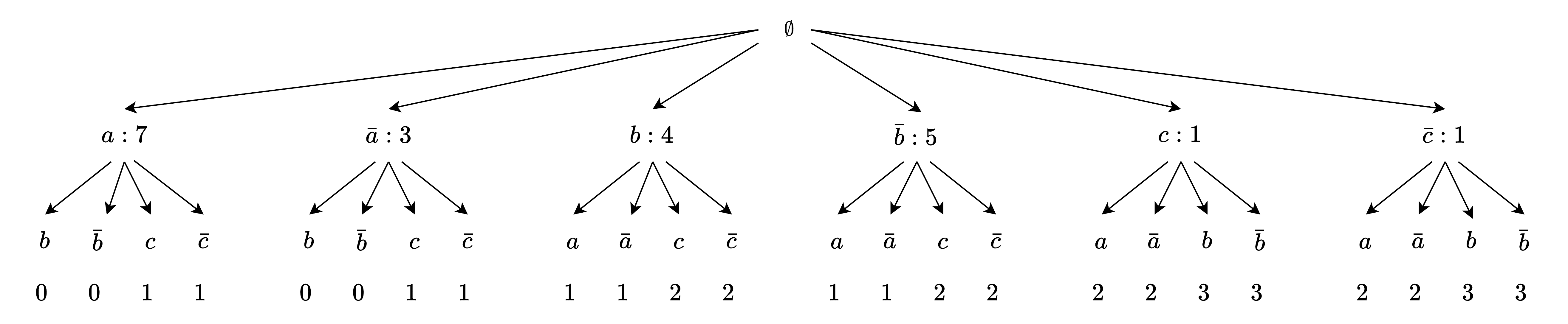}
\caption{Search tree for three features. The values at depth 1 represent node error before expansion; at depth 2, they reflect discrepancy costs.}
\label{fig:searchtree}
\end{figure*}

Overall, our work provides a unified and extensible framework for designing anytime exact decision tree algorithms, offering both theoretical guarantees and strong empirical performance. The CA-DL8.5 algorithm opens up new possibilities for combining optimality with real-time responsiveness in interpretable machine learning.
\section{Related Works and Background}
\label{sec:related}

\paragraph{Greedy approaches}
% Traditional tree-based algorithms like CART~\cite{breiman1984cart} and C4.5~\cite{quinlan2014c45} use greedy heuristics to efficiently construct decision trees through top-down splitting based on local informations. While scalable, these approaches cannot guarantee optimality. To address this limitation, researchers have explored optimization-based methods. Bertsimas and Dunn~\cite{bertsimas2017optimal} formulated optimal decision tree learning as a mixed-integer programming problem, while others leveraged SAT~\cite{narodytska2018learning}, MaxSAT~\cite{hu2020learning}, and constraint programming~\cite{verhaeghe2020learning}. Despite their theoretical guarantees, these approaches suffer from scalability issues on real world large datasets, especially for trees beyond shallow depths.
Traditional tree-based algorithms, such as CART~\cite{breiman1984cart} and C4.5~\cite{quinlan2014c45}, construct decision trees using greedy, top-down splits based on local information. These methods are efficient and scalable but cannot guarantee optimality. To address this limitation, several optimization-based approaches have been proposed: Bertsimas and Dunn~\cite{bertsimas2017optimal} formulated tree learning as a mixed-integer program, while others leveraged SAT~\cite{narodytska2018learning}, MaxSAT~\cite{hu2020learning}, and constraint programming~\cite{verhaeghe2020learning}. Although these methods provide theoretical guarantees, they struggle to scale to large datasets or deeper trees.

\paragraph{Dynamic Programming Approaches and DL8.5.}
Dynamic Programming (DP) methods~\cite{nijssen2007mining,aglin2020learning,demirovic2022murtree} improve the scalability of exact decision tree learning. DL8.5~\cite{aglin2020learning}, the foundation of our work, learns optimal decision trees on binary datasets $\mathcal{D}$ under minimum support and maximum depth constraints. A binary dataset $(\mathcal{D},\mathcal{C})$ consists of data  $\mathcal{D}\subseteq\prod_{f\in\mathcal{F}}\{f,\bar f\}$ over boolean features ${\cal F}$; $\mathcal{C}(x)$ indicates class labels for all $x\in {\cal D}$. For $\mathcal{S}\subseteq\mathcal{D}$ and feature $f$, define $\mathcal{S}(f)=\{x\in\mathcal{S}\mid f\in x\}$ and $\mathcal{S}(\bar f)$ analogously. A binary decision tree assigns features to internal nodes and labels edges with $f$ or $\bar f$, with each branch representing a root-to-leaf path. For branch $b$, $\mathcal{S}(b)=\bigcap_{f\in b}\mathcal{S}(f)$, and its classification error is $\lvert \mathcal{S}(b)\rvert-\max_{c\in\mathcal{C}}\lvert\mathcal{S}_c(b)\rvert$, where ${\cal C}$ indicates all class labels. 

DP-based algorithms such as DL8.5 perform a depth-first exploration of an \emph{OR-AND} search tree. As shown in Algorithm~{alg:dl85}, at each node, it selects a feature (OR node) and recursively explores the two branches $\{f,\bar f\}$ (AND nodes). The left branch is evaluated first (Line~\ref{alg:dl85:left}) by filtering examples not matching the feature, followed by the right branch (Line~\ref{alg:dl85:right}) with examples that do. This recursion progressively reduces the dataset and enforces the minimum support (Line~\ref{alg:dl85:minsup}) and maximum depth (Line~\ref{alg:dl85:maxdepth}) constraints. Subtree errors are computed and combined to obtain the current tree's error. Efficiency is enhanced through two mechanisms: upper-bound pruning (Lines~\ref{alg:dl85:ub1} and~\ref{alg:dl85:ub2}) discards unpromising branches, and a trie-based cache (Lines~\ref{alg:dl85:cache1} and~\ref{alg:dl85:cache2}) stores previously solved subproblems to avoid redundant computations.
% DP-based algorithms such as DL8.5 follow a depth-first strategy to explore an \emph{OR-AND} search tree. At each node, the algorithm iterates over the available features, recursively identifying the optimal subtree assuming a given feature is selected (representing the OR node). For each feature, it evaluates the two possible branches ${f, \bar{f}}$—first exploring the left subtree (Line~\ref{alg:dl85:left}) by filtering examples that do not match the feature, then the right subtree (Line~\ref{alg:dl85:right}) with examples that do (these represent the AND nodes). This recursive process reduces the problem size at each depth and respects constraints such as minimum support (Line~\ref{alg:dl85:minsup}) and maximum depth (Line~\ref{alg:dl85:maxdepth}). DL8.5 calculates the classification errors of subtrees and combines them to determine the overall error for the current tree. To enhance efficiency, the algorithm employs upper bound pruning (Lines~\ref{alg:dl85:ub1} and~\ref{alg:dl85:ub2}) to discard unpromising branches and uses a trie-based cache (Lines~\ref{alg:dl85:cache1} and~\ref{alg:dl85:cache2}) to store and reuse previously computed subtrees, significantly reducing redundant computations.

\begin{algorithm}[ht]
    \caption{DL8.5}
    \label{alg:dl85}
    \DontPrintSemicolon
    \small
    \SetKw{Break}{break}
    \SetKw{Continue}{continue}
    % \SetAlgoLined
    \SetKwInOut{Input}{Input}
    \SetKwInOut{Output}{Output}
    \SetKwRepeat{Do}{do}{while}
    \SetKwFunction{best}{best}
     \SetKwFunction{leaf}{leaf}
    \SetKwFunction{error}{error}
    \SetKwFunction{ub}{upperbound}
    
    \Input{$\mathcal{D},\: \mathtt{minsup},\: \mathtt{maxdepth}$}
    \Output{Best tree under the $\mathtt{maxdepth}$ and $\mathtt{minsup}$ constraints}

    $\mathbf{struct}\: \mathtt{Best}\{error: \mathtt{float}, ub: \mathtt{float}, tree: \mathtt{Tree}\}$\;
    $cache \gets Trie<branch,\: \mathtt{Best}>$\;
    $solution \gets \mathtt{DL85-Search}(\emptyset,\:, +\infty)$\;
    \Return{$solution$}\;

\SetKwFunction{proc}{$\mathtt{DL85-Search}$}
\SetKwProg{myproc}{Procedure}{}{}
\myproc{\proc{$b,\: ub$}} {

$e \gets \error(b)$\;
\If{$|b| = \mathtt{maxdepth}$ \textbf{or}  $ e = 0$}{
\label{alg:dl85:maxdepth}
        \Return{$\mathtt{Best} \{ e,\: ub,\: \leaf(b)\}$}\;

}

$node \gets cache.\mathtt{get}(b)$
\label{alg:dl85:cache1}

\If{$node \neq \emptyset$ \textbf{and} $ub \leq node.ub$} {
    \Return{$node$}\;
}

$(\tau,\: best\_error,\; child\_ub) \gets (\emptyset,\: +\infty,\: ub)$\;
\For{$f\: \mathbf{in}\: \mathcal{F}$ sorted by a heuristic}{
    \If{$|\mathcal{D}(b \cup \{f\})| < \mathtt{minsup}\: \mathbf{or}\: |\mathcal{D}(b \cup \{\bar f\})| < \mathtt{minsup}$ } {
    \label{alg:dl85:minsup}
        \Continue\;
    }

    $left \gets \mathtt{DL85-Search}(b \cup \{\bar f\},child\_ub)$\;
    \label{alg:dl85:left}
    \lIf{$left.tree = \emptyset$}{
    \label{alg:dl85:ub1}
      \Continue  
    }

     $right \gets \mathtt{DL85-Search}(b \cup \{f\},\:child\_ub - left.error)$\;
    \label{alg:dl85:right}
    \lIf{$right.tree = \emptyset$}{
    \label{alg:dl85:ub2}
      \Continue \;
    }

    $e \gets left.error + right.error$\;

    \If{$e < child\_ub$} {
        $best\_error \gets e$\;
        $child\_ub \gets e$\;
        $\tau \gets \mathtt{Tree}(left.tree, right.tree)$\;
    }
    \lIf{$e = 0$}{
        \Break
    }
    }
    $node \gets \mathtt{Best}\{best\_error,\: ub,\: \tau\}$\;
    $cache.save(b, node)$\;
    \label{alg:dl85:cache2}
    \Return{$node$}\;
}

\end{algorithm}

\paragraph{Anytime complete approaches}
As illustrated in Figure~\ref{fig:stuck_search}, the recursive depth-first exploration of DL8.5 prioritizes the leftmost branches of the search tree (solid lines), potentially delaying the exploration of other promising regions. Consider three binary features $a$, $b$, and $c$: the algorithm may fully explore the path $abc$ and its variations before considering alternative feature combinations. With many features and larger maximum depths, the search can remain trapped in early subtrees for extended periods, leaving large regions of the search space such as $ab\bar{c}$, $a\bar{b}$, and $\bar{a}$ (dashed lines) unexplored until late in the process.
% As illustrated in Figure~\ref{fig:stuck_search}, this recursive depth-first search exploration strategy causes the algorithm to prioritize the leftmost branches of the search tree (solid lines), potentially delaying the exploration of other promising regions. Consider a scenario with three binary features $a$, $b$, and $c$. The search might become fixated on exploring the path $abc$ and its variants in depth before considering other feature combinations. When dealing with many features and larger maximum depths, the algorithm can remain trapped in these early subtrees for extended periods. Consequently, large regions of the search space—such as the alternative paths $ab\bar{c}$, $a\bar{b}$, $\bar{a}$ (dashed lines in the figure)—remain unexplored until late in the search process.
This behavior introduces a key weakness: if interrupted, the algorithm typically produces an incomplete, unbalanced decision tree, that can be of lower quality than one built by simple greedy heuristics. To address this limitation and improve anytime performance, we propose integrating a complete anytime beam search strategy into DL8.5, preserving optimality guarantees while producing high-quality solutions throughout execution.

% This behavior creates a critical weakness: if the algorithm is interrupted before completion, the intermediate solution is typically an unbalanced decision tree, incomplete, and can be of even lower quality than trees produced by simple greedy heuristics. To overcome this limitation and improve anytime behavior, we propose integrating a complete anytime beam search strategy into DL8.5, resulting in an algorithm that maintains optimality guarantees while providing high-quality solutions at any point during execution.
Recent work has explored improving the anytime behavior of exact decision tree algorithms. Kiossou et al.~\cite{kiossou2022time} introduced LDS-DL8.5, which applies iterative Limited Discrepancy Search (LDS) to prioritize solutions close to a heuristic baseline tree. A discrepancy corresponds to selecting a feature different from the one chosen by a greedy algorithm (e.g., the feature with maximum information gain). By gradually increasing the allowed number of discrepancies along a root-to-leaf path, the algorithm transitions from the greedy tree (zero discrepancies) toward an exhaustive search. This approach makes the algorithm more efficient in discovering the best tree. Early discovery of good decision trees helps prune the search space using the branch-and-bound techniques of DL8.5. In the worst case, the best tree found coincides with the greedy tree.

% Several recent approaches have specifically aimed to improve the anytime behavior of complete algorithms for decision trees. \cite{kiossou2022time} proposed LDS-DL8.5, which incorporates an iterative limited discrepancy search to prioritize solutions with small deviations from an initial heuristic tree.
% A discrepancy is defined as a deviation from a default heuristic. In the context of decision trees, it refers to exploring alternative features that differ from the one selected by a greedy algorithm (for example, the feature with the highest information gain). By gradually increasing the discrepancy limit—the maximum number of allowed discrepancies along a complete path in the decision tree—the search process transitions iteratively. Initially, with a discrepancy budget of zero, it discovers the same tree as the greedy algorithm. As the limit increases, the search progressively moves toward a more exhaustive exploration.
% This approach makes the algorithm more efficient in discovering the best tree. Early discovery of good decision trees helps prune the search space using the branch-and-bound techniques of DL8.5. This also allows the user to stop the search earlier while still obtaining a good tree. In the worst case, the tree found will be the same as the one identified by a greedy strategy.
% \psc{Not very clear to me $\rightarrow$}
% MurTree~\cite{demirovic2022murtree} attempts to address this issue by dynamically selecting subtrees to branch on, but it still cannot guarantee good solution quality  when interrupted prematurely.
The Blossom algorithm, introduced in~\cite{demirovic2023blossom}, employs a different search approach to enhance the anytime behavior of the search. While it also utilizes depth-first search, it proceeds layer by layer within the tree, always expanding the non-expanded node that is closest to the root. Similar to LDS-DL8.5, the first solution found in the leftmost leaf node of the depth-first search exploration corresponds to the decision tree that would be identified by a purely greedy strategy. This characteristic makes it an improvement in terms of anytime behavior compared to DL8.5.

However, like DL8.5, the Blossom algorithm can also suffer from poor diversification of the search space, particularly concerning features of nodes close to the root. These features are reconsidered last during backtracking, despite being selected first and somewhat blindly when limited information was available for making an informed decision. To the best of our knowledge, no experiments have yet been conducted to compare Blossom and LDS-DL8.5.

\paragraph{Anytime incomplete approaches}
The Top-$k$ method, introduced in~\cite{blanc2024harnessing} is an extension of DL8.5 that aims at improving its scalability. It shares many similarities with LDS-DL8.5. Specifically, it heuristically restricts the set of features tried at each OR-node to the Top-$k$ features, as determined by some ranking heuristic. The search space to explore is thus voluntarily limited and therefore this method can be considered as a trade-off between the pure greedy C4.5 and the complete search of DL8.5.

Other hybrid methods like LGDT~\cite{kiossou2024efficient} balance greedy efficiency with limited depth lookahead based on exact method to improve solution quality without the full computational cost of exact methods.

% \psc{shall we also mention the local search approach of Bertsimas}

\section{Complete Anytime Beam Search DL8.5 (CADL8.5)}
\label{sec:cadl85}
Beam search is an informed search algorithm that explores a space by keeping only the most promising nodes at each level. While it improves upon breadth-first and depth-first search through heuristic pruning, overly aggressive pruning can cause it to miss solutions. To overcome this, Zhang~\cite{zhang1998complete} introduced Complete Anytime Beam Search (CABS), based on the principle of \emph{iterative weakening}~\cite{provost1993iterative}. 
% Beam search is an informed search algorithm that explores a search space by maintaining only a number of the most promising nodes at each exploration level. While standard beam search improves upon best-first search (BFS) and depth-first search (DFS) by using heuristic pruning to limit exploration, it may fail to find solutions if its pruning is too aggressive. To address this limitation, Zhang~\cite{zhang1998complete} proposed Complete Any-time Beam Search (CABS), which builds upon the concept of iterative weakening~\cite{provost1993iterative}.
CABS performs successive beam search iterations with progressively relaxed pruning constraints. It starts with strict pruning, quickly identifying initial solutions, and then gradually weakens the pruning rules to explore larger portions of the search space. This process enables the algorithm to discover solutions that would have been pruned in earlier iterations.
% More exactly CABS performs multiple beam search iterations with progressively relaxed pruning constraints. The search begins with strict pruning rules that focus exploration on the most promising paths, enabling quick identification of initial solutions. If these solutions do not meet the desired criteria, the algorithm systematically weakens the pruning rules and performs additional search iterations. Each subsequent iteration explores a broader portion of the state space, allowing the algorithm to discover potentially better solutions that might have been pruned in earlier iterations.

This iterative relaxation approach gives CABS two valuable properties: (1) it provides anytime behavior by quickly finding initial solutions that improve over time, and (2) it ensures completeness by gradually reducing pruning constraints until an optimal solution is found. These properties make CBS particularly well-suited for complex optimization problems like decision tree learning, where balancing solution quality with computational efficiency is important.

In this work we propose Complete Anytime DL8.5 (CADL8.5), which  adapts the original DL8.5 algorithm by including CABS ideas.
It is a generic framework that guarantees high-quality solutions even when terminated early. 
By integrating principles from CABS, CADL8.5 guides tree construction using adaptive pruning rules. 
These rules are relaxed over multiple iterations, allowing the algorithm to prioritize promising regions initially while progressively exploring more diverse parts of the solution space.

Algorithm~\ref{alg:cadl85} presents the structure of CADL8.5. 
In addition to the standard inputs required by DL8.5 (dataset $\mathcal{D}$, minimum support $\mathtt{minsup}$, and maximum depth $\mathtt{maxdepth}$), CADL8.5 requires one additional parameter: $\mathtt{r}$, the rule or the set of rules guiding the search space exploration. This allows the algorithm to balance between quickly finding initial solutions and ensuring optimality through complete exploration.

% \begin{algorithm}[ht]
%     \caption{Complete Beam Search}
%     \label{alg:cbs}
%     \DontPrintSemicolon
%     \SetKw{Break}{break}
%     \SetKw{Continue}{continue}
%     % \SetAlgoLined
%     \SetKwInOut{Input}{Input}
%     \SetKwInOut{Output}{Output}
%     \SetKwRepeat{Do}{do}{while}
%     \Input{\textit{problem}, \textit{R}, $\alpha$}
%     \Output{\textit{goal}}
%     $goal \gets \emptyset$\;
%     \Do{goal is not desired \& R not empty }{
%         $goal \gets \mathtt{BeamSearch}(problem, R, \alpha$)\;
%         weaken rules in $R$\;
%          \label{cbs:weaken}
%     }

% \;
% \SetKwFunction{proc}{$\mathtt{BeamSearch}$}
% \SetKwProg{myproc}{Procedure}{}{}
% \myproc{\proc{$problem,\: R,\: \alpha$}} {

% $nodes \gets \{problem\}$\;

% \While{$nodes \neq \emptyset$}{

% pick and remove a node $n$ from $nodes$\;

% \lIf{$n$ is desired goal}{
% \Break
% }

% \lIf{$n$ is a goal \& $\mathtt{cost}(n) < \alpha$}{
% $\alpha \gets \mathtt{cost}(n)$
% }

% \For{$child$ in $\mathtt{children}(n)$ }{
% \label{cbs:loopstart}
% \lIf{$\mathtt{cost}(child) \geq \alpha$ \textbf{or} pruned by a rule in $R$ }{
% \Continue
% }
% $nodes \gets nodes \cup \{child\}$
% }
% \label{cbs:loopend}
% }
% }

% \end{algorithm}

\begin{table}[ht]
\scriptsize
\centering
\setlength{\tabcolsep}{3pt}
\renewcommand{\arraystretch}{1.15}
\begin{tabular}{@{}lll@{}}
\toprule
\textbf{Rule} & \textbf{State \& Constraint} & \textbf{Key Functions} \\
\midrule
\textbf{Purity} &
\begin{tabular}[t]{@{}l@{}}
$\mathtt{T}\{\mathtt{purity: float}\}$ \\
$\mathtt{R}\{\mathtt{min\_purity: float}\}$
\end{tabular} &
\begin{tabular}[t]{@{}l@{}}
$\mathtt{update}: \mathtt{purity} = 1 - \frac{\mathtt{ctx.e}}{|\mathtt{ctx.S}|}$ \\
$\mathtt{prune}: \mathtt{purity} \geq \mathtt{min\_purity}$ \\
$\mathtt{relax}: \mathtt{min\_purity} \mathbin{{+}{=}} \delta$
\end{tabular} \\
\midrule
\textbf{Gain} &
\begin{tabular}[t]{@{}l@{}}
$\mathtt{T}\{\mathtt{cum\_gap: float}\}$ \\
$\mathtt{R}\{\mathtt{max\_gap: float}\}$
\end{tabular} &
\begin{tabular}[t]{@{}l@{}}
$\mathtt{update}: \mathtt{cum\_gap} \mathbin{{+}{=}} \mathtt{best\_gain} - \mathtt{gain}$ \\
$\mathtt{prune}: \mathtt{cum\_gap} > \mathtt{max\_gap}$ \\
$\mathtt{relax}: \mathtt{max\_gap} \mathbin{{+}{=}} \delta$
\end{tabular} \\
\midrule
\textbf{Discrepancy} &
\begin{tabular}[t]{@{}l@{}}
$\mathtt{T}\{\mathtt{tot\_discr: int}\}$ \\
$\mathtt{R}\{\mathtt{max\_discr: int}\}$
\end{tabular} &
\begin{tabular}[t]{@{}l@{}}
$\mathtt{update}: \mathtt{tot\_discr} \mathbin{{+}{=}} \mathtt{ctx.i}$ \\
$\mathtt{prune}: \mathtt{tot\_discr} \geq \mathtt{max\_discr}$ \\
$\mathtt{relax}: \mathtt{max\_discr} \mathbin{{+}{=}} \delta$
\end{tabular} \\
\midrule
\textbf{Top-$k$} &
\begin{tabular}[t]{@{}l@{}}
$\mathtt{T}\{\mathtt{feat\_idx: int}\}$ \\
$\mathtt{R}\{\mathtt{k: int}\}$
\end{tabular} &
\begin{tabular}[t]{@{}l@{}}
$\mathtt{update}: \mathtt{feat\_idx} = \mathtt{ctx.i}$ \\
$\mathtt{prune}: \mathtt{feat\_idx} \geq \mathtt{k}$ \\
$\mathtt{relax}: \mathtt{k} \mathbin{{+}{=}} \delta$
\end{tabular} \\
\bottomrule
\end{tabular}
\caption{Rule definitions for CADL8.5}
\label{tab:rules}
\end{table}

\begin{algorithm}[!h]
\caption{Complete Anytime DL8.5 (CADL8.5)}
\label{alg:cadl85}
\DontPrintSemicolon
\scriptsize
%\SetAlgoLined
\SetKwInOut{Input}{Input}
\SetKwInOut{Output}{Output}
\SetKw{Break}{break}
\SetKw{Continue}{continue}
\SetKwRepeat{Do}{do}{while}
\SetKwFunction{Best}{Best}
\SetKwFunction{Leaf}{leaf}
\SetKwFunction{Error}{error}
\SetKwFunction{Search}{BeamDL8.5}
\SetKwFunction{Prune}{prune}
\SetKwFunction{Relax}{relax}
\SetKwFunction{Update}{update}
\SetKwFunction{InitialState}{initial\_state}
\SetKwFunction{TerminalState}{terminal\_state}

\Input{$\mathcal{D}$, $\mathtt{rule}$, minsup, maxdepth}
\Output{Best tree satisfying minsup and maxdepth}

\textbf{Struct} $\mathtt{Best}\{ \mathtt{error}:\mathrm{float}, \mathtt{ub}:\mathrm{float}, \mathtt{tree}:\mathrm{Tree}, \mathtt{state}: \mathtt{T} \}$\;
$cache \gets \texttt{Trie}<\texttt{branch}, \texttt{Best}>$\;
$ub \gets +\infty$\;

$context_0 \gets$ \{ i: 0, e: $\Error(\emptyset)$, s: $|\mathcal{D}|$ \}\;
$state_0 \gets \InitialState(context_0)$\;
\label{alg:cadl85:initstate}

\Do{$sol$ is not optimal \textbf{or} $\mathtt{rule}$ is relaxable}{
\label{alg:cadl85:outerloop1}
    $sol \gets \Search(\emptyset, ub, context_0, state_0)$\;
    $ub \gets sol.error$\;
    $\mathtt{rule} \gets \Relax(\mathtt{rule})$\;
    $state_0 \gets sol.state$\;
}
\label{alg:cadl85:outerloop2}
\Return{$sol.tree$}

\BlankLine
\SetKwProg{Procedure}{Procedure}{}{}
\Procedure{$\Search(b, ub, context_p, state_p)$}{

    $e \gets \Error(b)$\;
    \If{$|b| = \texttt{maxdepth}$ \textbf{or} $e = 0$}{
    \label{alg:cadl85:terminal}
        \Return $\Best\{ e, ub, \Leaf(b), \TerminalState(state_p) \}$\;
    }

    \If{time limit reached \textbf{or} $\Prune(state_p, \mathtt{rule})$}{
    \label{alg:cadl85:costpruning}
        \Return $\Best\{ e, ub, \Leaf(b), state_p \}$\;
    }

    $node \gets cache.\texttt{get}(b)$\;
    \If{$node \neq \emptyset$ \textbf{and} $ub \leq node.ub$ \textbf{and} $\neg \Prune(node.state, \mathtt{rule})$}{
        \label{alg:cadl85:cachereturn}
        \Return{$node$}\;
    }

    $\tau \gets \emptyset$, $child\_ub \gets ub$, $optimal \gets$ \texttt{true}\;

    \ForEach{$(i, f)$ \textbf{in} $\mathcal{F}$ sorted by heuristic}{
        \If{$|\mathcal{D}(b \cup \{f\})| < \texttt{minsup}$ \textbf{or} $|\mathcal{D}(b \cup \{\bar{f}\})| < \texttt{minsup}$}{
            \Continue
        }

        $context_l \gets$ \{ i: $i$, e: $\Error(b \cup \{\bar{f}\})$, s: $|\mathcal{D}(b \cup \{\bar{f}\})|$ \}\;
        $state_l \gets \Update(state_p, context_l)$\;
        \label{alg:cadl85:cost1}
        $l \gets \Search(b \cup \{\bar{f}\}, child\_ub, context_l, state_l)$\;
        \If{$l.tree = \emptyset$}{ \Continue }

        $context_r \gets$ \{ i: $i$, e: $\Error(b \cup \{f\})$, s: $|\mathcal{D}(b \cup \{f\})|$ \}\;
        $state_r \gets \Update(state_p, context_r)$\;
        $r \gets \Search(b \cup \{f\}, child\_ub - l.error, context_r, state_r)$\;
        \If{$r.tree = \emptyset$}{ \Continue }

        $e \gets l.error + r.error$\;

        \If{$\Prune(l.state, \mathtt{rule})$ \textbf{or} $\Prune(r.state, \mathtt{rule})$}{
        \label{alg:cadl85:notoptimal}
            $optimal \gets$ \texttt{false}\;
        }

        \If{$e < child\_ub$}{
            $child\_ub \gets e$\;
            $\tau \gets \texttt{Tree}(l.tree, r.tree)$\;
            $context_p.error \gets e$\;
            $state_p \gets \Update(state_p, context_p)$\;
            \label{alg:cadl85:cost2}
        }

        \If{$e = 0$}{ \Break }
    }

    \If{$optimal$}{ $state_p \gets \TerminalState(state_p)$ }
    \label{alg:cadl85:caching1}
    $node \gets \Best\{ child\_ub, ub, \tau, state_p \}$\;
    $cache.\texttt{save}(b, node)$\;
    \label{alg:cadl85:caching2}
    \Return{$node$}\;
}
\end{algorithm}

A key difference between DL8.5 and CADL8.5 is the introduction of an outer loop (Lines~\ref{alg:cadl85:outerloop1}–\ref{alg:cadl85:outerloop2}) that repeatedly invokes the $\mathtt{BeamDL8.5}$ search procedure while progressively relaxing constraints. The algorithm begins with strict pruning rules to rapidly identify a feasible solution that establishes an initial upper bound on the classification error. After each iteration, the rules are weakened using the $\mathtt{relax}$ function, expanding the search space. This process continues until one of three conditions is met: a perfect tree is found (zero error), no further rule relaxation is possible, or the allocated time budget is exhausted.

To implement this strategy, CADL8.5 introduces two generic types: $\mathtt{T}$ and $\mathtt{R}$. Type $\mathtt{T}$ encapsulates state information required to apply pruning decisions, while $\mathtt{R}$ stores the parameters that define the current rule constraints. These types are manipulated through five core functions:
\begin{itemize}
    \item $\mathtt{update}(t: \mathtt{T}, c: \mathtt{Context}) \rightarrow \mathtt{T}$: modifies a node's state based on its context, including error metrics, dataset size, and feature selection information.
    \item $\mathtt{prune}(t: \mathtt{T}, r: \mathtt{R}) \rightarrow \mathtt{boolean}$: evaluates whether a node should be pruned based on its current state and rule parameters.
    \item $\mathtt{relax}(r: \mathtt{R}) \rightarrow \mathtt{R}$: incrementally weakens rule constraints to permit wider exploration in subsequent iterations.
    \item $\mathtt{terminal\_state}(t: \mathtt{T}) \rightarrow \mathtt{T}$: generates a special state marking a node as fully explored and exempt from future pruning.
    \item $\mathtt{initial\_state}(c: \mathtt{Context}) \rightarrow \mathtt{T}$: creates the initial state for the root node based on its context.
\end{itemize}

The search process begins by constructing an initial context $c_0$ at the root, incorporating the dataset's error measure and size. This context is passed to $\mathtt{initial\_state}$ to generate the root's initial state $t_0$ (Line~\ref{alg:cadl85:initstate}), which is then used alongside the initial upper bound to start the first $\mathtt{BeamDL8.5}$ search iteration. 

At the root of the search space, the initial context $c_0$ is constructed using the root error and dataset size. This context is then passed to $\mathtt{initial\_state}$ to produce the initial state $t_0$ (Line~\ref{alg:cadl85:initstate}), which is used to start the first call to $\mathtt{BeamDL8.5}$ along with the initial upper bound.

During the $\mathtt{BeamDL8.5}$ procedure, each node's state is evaluated against the current rule configuration before expansion. The $\mathtt{update}$ function computes an updated state $t$ based on the node's context (Lines~\ref{alg:cadl85:cost1} and~\ref{alg:cadl85:cost2}). This state is then evaluated using the $\mathtt{prune}$ predicate (Line~\ref{alg:cadl85:costpruning}) to determine if the node violates any active constraints. If pruning conditions are met, the node is not expanded further and is returned as-is. 

Nodes are also not expanded when they reach terminal conditions: either the maximum tree depth is attained or perfect classification (zero error) is achieved. In these cases, the node is explicitly marked as fully explored by applying the $\mathtt{terminal\_state}$ function (Line~\ref{alg:cadl85:terminal}). This ensures that terminal nodes are treated as optimal in future caching operations and pruning decisions, preventing unnecessary recomputation of already optimal subtrees. 

CADL8.5 extends DL8.5's caching strategy to avoid redundant computations across iterations. Each branch $b$ is associated with a $\mathtt{Best}$ object that stores the optimal subtree, its error bound, and its state. Before expanding any node, the cache is queried for previous results. If a compatible cached entry exists—one whose upper bound is consistent with current requirements and not pruned under the current rule set—the cached result is immediately reused (Line~\ref{alg:cadl85:cachereturn}). Additionally, nodes are marked as fully explored only when all their children have been optimally processed (Lines~\ref{alg:cadl85:caching1}–\ref{alg:cadl85:caching2}).

To control the exploration cost of nodes in CADL8.5, we implement four different pruning strategies: the \textit{Purity} rule, the \textit{Gain} rule, the \textit{Discrepancy} rule, and the \textit{Top-$k$} rule. Table~\ref{tab:rules} summarizes these rules using the generic type structures introduced earlier.

\begin{table*}[ht]
\centering
\scriptsize
\begin{minipage}{0.48\textwidth}
\centering
\begin{tabular}{llrrrrrr}
\toprule
 & & \multicolumn{6}{c}{Runtime (s)} \\
Approach & Sub & 15 & 30 & 60 & 120 & 240 & 300 \\
\midrule
\multirow{1}{*}{C4.5} & -- & 64.3 & 64.3 & 64.3 & 64.3 & 64.3 & 64.3 \\
\midrule
\multirow{1}{*}{Top-$3$} & -- & 46.3 & 45.0 & 44.4 & 44.1 & 44.0 & 44.0 \\
\midrule
\multirow{1}{*}{Top-$5$} & -- & 37.5 & 35.2 & 34.3 & 33.8 & 33.6 & 33.5 \\
\midrule
\multirow{1}{*}{DL8.5} & -- & 46.6 & 40.4 & 34.6 & 31.5 & 29.5 & 28.8 \\
\midrule
\multirow{1}{*}{Blossom} & -- & \textbf{27.1} & 24.7 & 19.6 & 14.5 & 9.6 & 8.5 \\
\midrule
\multirow{1}{*}{CA-Purity} & -- & 48.8 & 42.5 & 36.8 & 31.6 & 27.7 & 26.7 \\
\midrule
\multirow{3}{*}{CA-Gain} & exponential & 43.7 & 36.3 & 32.2 & 29.2 & 26.4 & 24.7 \\
& luby & 42.8 & 35.9 & 31.5 & 25.5 & 22.0 & 21.2 \\
& monotonic & 43.2 & 36.6 & 32.3 & 26.2 & 22.5 & 21.5 \\
\midrule
\multirow{3}{*}{CA-Discrepancy} & exponential & 29.1 & 23.2 & 18.6 & 15.8 & 13.0 & 11.4 \\
& luby & 27.2 & \textbf{20.7} & \textbf{16.7} & 13.9 & 9.2 & 7.8 \\
& monotonic & 27.4 & 20.8 & 16.7 & 14.0 & 8.8 & \textbf{7.4} \\
\midrule
\multirow{3}{*}{CA-Top-$k$} & exponential & 34.9 & 30.6 & 24.9 & 20.0 & 17.5 & 16.8 \\
& luby & 32.7 & 25.4 & 20.5 & 17.3 & 14.0 & 12.0 \\
& monotonic & 31.1 & 23.5 & 19.0 & 16.3 & 13.4 & 11.3 \\
\midrule
\multirow{3}{*}{CA-Top-$k^*$} & exponential & 34.4 & 27.2 & 23.3 & 17.0 & 12.7 & 11.4 \\
& luby & 31.0 & 25.4 & 18.0 & \textbf{12.2} & \textbf{8.5} & 7.7 \\
& monotonic & 31.2 & 25.6 & 18.4 & 12.2 & 8.6 & 7.7 \\
\bottomrule
\end{tabular}
\caption{Average primal integral on depth $6$}
\label{tab:d6_p}
\end{minipage}
\hfill
\begin{minipage}{0.48\textwidth}
\centering
\begin{tabular}{llrrrrrr}
\toprule
 & & \multicolumn{6}{c}{Runtime (s)} \\
Approach & Sub & 15 & 30 & 60 & 120 & 240 & 300 \\
\midrule
\multirow{1}{*}{C4.5} & -- & 69.3 & 69.3 & 69.3 & 69.3 & 69.3 & 69.3 \\
\midrule
\multirow{1}{*}{Top-$3$} & -- & 52.9 & 51.7 & 51.1 & 50.8 & 50.7 & 50.6 \\
\midrule
\multirow{1}{*}{Top-$5$} & -- & 46.8 & 41.0 & 38.5 & 37.2 & 36.6 & 36.5 \\
\midrule
\multirow{1}{*}{DL8.5} & -- & 58.3 & 52.0 & 47.2 & 43.9 & 40.4 & 39.5 \\
\midrule
\multirow{1}{*}{Blossom} & -- & 37.1 & 30.5 & 24.7 & 21.5 & 18.6 & 17.2 \\
\midrule
\multirow{1}{*}{CA-Purity} & -- & 55.9 & 50.0 & 45.7 & 43.2 & 40.7 & 39.7 \\
\midrule
\multirow{3}{*}{CA-Gain} & exponential & 50.5 & 45.3 & 39.1 & 35.5 & 33.0 & 32.5 \\
& luby & 54.7 & 51.4 & 45.6 & 41.7 & 38.0 & 36.9 \\
& monotonic & 55.1 & 51.6 & 46.2 & 41.9 & 39.3 & 38.5 \\
\midrule
\multirow{3}{*}{CA-Discrepancy} & exponential & 33.0 & 28.3 & 25.0 & 22.8 & 19.3 & 17.8 \\
& luby & \textbf{31.3} & 26.9 & 22.8 & 19.0 & 15.1 & \textbf{14.0} \\
& monotonic & 31.5 & \textbf{26.5} & \textbf{22.5} & \textbf{19.0} & \textbf{15.1} & 14.0 \\
\midrule
\multirow{3}{*}{CA-Top-$k$} & exponential & 41.8 & 34.5 & 28.5 & 25.1 & 22.4 & 21.1 \\
& luby & 39.4 & 31.2 & 25.3 & 22.1 & 19.1 & 18.0 \\
& monotonic & 37.7 & 29.7 & 24.2 & 20.8 & 17.6 & 16.6 \\
\midrule
\multirow{3}{*}{CA-Top-$k^*$} & exponential & 41.8 & 36.7 & 33.9 & 31.3 & 27.7 & 26.6 \\
& luby & 39.6 & 34.7 & 30.7 & 26.7 & 23.4 & 22.3 \\
& monotonic & 41.8 & 35.8 & 31.3 & 26.9 & 23.8 & 22.6 \\
\bottomrule
\end{tabular}
\caption{Average primal integral on depth $7$}
\label{tab:d7_p}
\end{minipage}
\end{table*}

\subsection{Purity rule}
The purity rule defines a minimum purity threshold for the tree stored in $\mathtt{R}$. A node expansion is stopped when its purity meets or exceeds this threshold. Weakening the rule involves incrementally increasing the threshold by a value $\delta$ until it reaches a maximum of $1.0$. If a node's purity remains below the current threshold (\(\mathtt{prune}\)), the search continues along that branch until the maximum depth is reached, during which purity may improve. Without a depth constraint, this strategy would attempt to construct a perfect decision tree. The purity of a branch \(b\) is defined as
\[
    purity(b) = 1 - \frac{error(b)}{|\mathcal{S}(b)|}.
\]
% where $error(b)$ is the number of misclassified samples in branch $b$, and $|\mathcal{S}(b)|$ is the number of samples in that branch.  
For example, in the search tree of Figure~\ref{fig:searchtree}, assume $10$ examples fall into each branch at depth $1$, with $purity(a) = 0.3$ and $purity(\bar a) = 0.7$.  
With a threshold of $0.5$, branch $a$ is expanded, whereas $\bar a$ is not, since it is pure enough.

% The purity rule defines a minimum purity threshold for the tree stored in $\mathtt{R}$, meaning that the expansion of a node is stopped when its purity reaches or exceeds this threshold. . Weakening the purity rule involves increasing the minimum purity threshold by a $\delta$ until the maximum of 1.0 is reached. If a node’s purity is below the expected threshold ($\mathtt{prune})$, its exploration continues until the maximum depth is reached, while its purity improves. Without the depth constraint, this approach would attempt to find the perfect tree, resulting in perfect classification.
% The purity of a branch is defined as:

% \begin{equation}
%     purity(b) = 1 - \frac{error(b)}{|\mathcal{S}(b)|},
% \end{equation}
% where $error(b)$ is the branch $b$ error, and $|\mathcal{S}(b)|$ is the number of samples in the branch. For example, in the search tree shown in Figure~\ref{fig:searchtree}, assuming $10$ examples fall into each branch at depth $1$, $purity(a) = 0.3$ and $purity(\bar a) = 0.7$. With a threshold of $0.5$, branch $a$ will be expanded, while $\bar a$ will not be, since it is pure enough. Weakening the purity rule involves increasing the minimum purity threshold until the maximum of $1.0$ is reached.

\subsection{Gain rule}
The gain rule restricts feature expansion using an information gain threshold. 
At a node, let $\tau^*$ be the highest gain and $\tau(f)$ the gain of a feature $f$. 
The local gap is $\tau^*-\tau(f)$, and each feature maintains a cumulative gap along the path from the root:

\[
\mathtt{cum\_gap} = \mathtt{cum\_gap}_{\text{parent}} + (\tau^* - \tau(f)).
\]

A feature is expanded only if $\mathtt{cum\_gap} \leq \mathtt{max\_gap}$. 
Setting $\mathtt{max\_gap}=0$ yields a greedy strategy similar to C4.5, 
while larger values allow broader exploration. 
The cumulative constraint naturally tightens at deeper levels, focusing the search and avoiding excessive exploration of suboptimal branches.
% The gain rule defines an information gain threshold to restrict the set of features considered for expansion at each node. For a given node, let $\tau^*$ denote the highest information gain among all candidate features. For any feature $f$ with gain $\tau(f)$, the local gap is defined as $\tau^* - \tau(f)$.
% Each node maintains a cumulative gap  $\mathtt{cum\_gap}$ for each feature $f$, computed recursively as the sum of the local gaps along the path from the root to the current node:
% \[
% \mathtt{cum\_gap} = \mathtt{cum\_gap}_{\text{parent}} + (\tau^* - \tau(f)).
% \]
% Only features for which $\mathtt{cum\_gap} \leq \mathtt{max\_gap}$ are considered for expansion at a node.
% Setting $\mathtt{max\_gap} = 0$ results in a purely greedy expansion, where only the best feature at each node is selected, similar to C4.5. Increasing $\mathtt{max\_gap}$ weakens the rule and allows the exploration of more diverse features. The cumulative nature of the constraint enforces a stricter selection in deeper parts of the tree, preventing getting stuck in lower part of the search space.

\subsection{Discrepancy rule}
The Discrepancy rule employs the same principle of Limited Discrepancy Search (LDS) as in the LDS-DL8.5 algorithm~\cite{kiossou2022time}, to control deviations from a preferred exploration order. Each node tracks the total discrepancy $\mathtt{tot\_discr}$ accumulated from the root, where each feature is assigned an index $i$ based on its position in the candidate list. The discrepancy of a path thus reflects how many times the search deviated from the leftmost option.

At each node, only features whose associated cost does not exceed the threshold $\mathtt{max\_discr}$ are considered. For example, exploring only the leftmost feature at each split (with $\mathtt{max\_discr} = 0$) results in a greedy tree. Increasing the discrepancy budget expands the search space and enables exploring other parts of the space search.

As illustrated in Figure~\ref{fig:searchtree}, if feature $A$ is explored first, then $cost(a) = cost(\bar a) = 0$. Choosing $B$ instead at the same level requires $cost(b) = cost(\bar b) = 1$. Similarly, deeper paths such as $ba$ and $b\bar{a}$ have $cost = 1$ since $A$ is the first successor of $b$, and $cost = 2$ for branches like $bc$ since $C$ is the second.

\subsection{Top-k rule}
The \emph{Top-\(k\) rule} controls the breadth of the search by limiting the number of features explored at each node. It considers only the \(k\) best candidates, where the position of a feature in the sorted list determines its index \(\mathtt{ctx.i}\). When \(k = 1\), the algorithm behaves greedily, producing trees similar to CART or C4.5 depending on the heuristic used. 
% As \(k\) increases, more features are evaluated per node, expanding the search space and allowing corrections to early decisions.
As \(k\) increases, more features are evaluated per node, expanding the search space and allowing corrections to early decisions.
We also propose a new variant, denoted as Top-$k$* in the results, where the beam width $k$ is halved at each level of the tree, with a minimum value of one. This allows to reduce the time spent in the deeper parts of the search space in early iterations. 

Unlike the Discrepancy rule, Top-\(k\) and Top-\(k^*\) rules do not accumulate costs across the tree. The feature index is local to the node and does not depend on the path. A node is pruned if its feature index exceeds the current threshold \(k\), unless marked as terminal. The relaxation function increments \(k\), progressively weakening the pruning condition.
The state tracks the index of the selected feature (\texttt{feat\_idx}), and pruning is bypassed when this index is set to \(\infty\), used as a sentinel for terminal nodes.

\section{Results}
To evaluate the performance of CADL8.5, we conducted a series of experiments. This section presents the results. We begin by analyzing the anytime behavior of CADL8.5 using the previously mentioned rules, followed by a comparison of its performance in finding optimal solutions. All experiments were conducted on 25 datasets from CP4IM, with a minimum support threshold of 1. The algorithms were executed on a server equipped with an Intel Xeon Platinum 8160 CPU and 320 GB of RAM, running Rocky Linux version 8.4. For comparison, we include a scikit-learn implementation of C4.5\footnote{\url{https://scikit-learn.org/}}, DL8.5 and Blossom implementations.
We compare CADL8.5 to the other algorithms using the \emph{average primal integral}, as introduced in~\cite{BERTHOLD2013611} to measure the anytime behavior of optimization solvers. The primal integral aims to measure the progress of an algorithm's primal bound convergence toward the optimal (or best known) solution over the entire solving time. It is based on the \textit{primal function} \( p(t) \), which represents the gap between the current solution \( x(t) \) at time \( t \) and the optimal or best known solution \( x_{\text{opt}} \). The \textit{primal gap} of a solution \( x(t) \) is defined as
\[
\gamma(x(t)) = \frac{|x(t) - x_{\text{opt}}|}{|x(t)|}.
\]
The function \( p(t) \) equals 1 if no feasible solution has been found by time \( t \), and \( \gamma(x(t)) \) otherwise. The function \( p(t) \) is a step function that changes whenever a new feasible solution is found. It is monotonically decreasing and becomes zero once the optimal solution is reached. The primal integral \( P(T) \) is defined as the integral of the primal gap function \( p(t) \) over the time horizon \( T \):
\[
P(T) = \int_0^T p(t) \, dt = \sum_{i=1}^{n} p(t_{i-1}) \cdot (t_i - t_{i-1}),
\]
where each \( t_i \) denotes a time point at which a new incumbent solution is found.
The primal integral encourages the early discovery of high-quality solutions. If a better solution is found at the same time, \( P(t_{\max}) \) decreases. Similarly, if the same solution is found earlier, \( P(t_{\max}) \) also decreases. The ratio \( P(t_{\max}) / t_{\max} \) can be interpreted as the average quality of the solution during the search process. A smaller value indicates a higher expected quality of the current solution if the algorithm is interrupted at an arbitrary point in time. 

% ---------------------

Tables~\ref{tab:d6_p} and~\ref{tab:d7_p} report the evolution of the average primal integral across various time budgets (from $15$ to $300$ seconds) for tree depths $6$ and $7$. To ensure meaningful comparisons, we exclude datasets where DL8.5 finds the optimal solution in under $1$ second. For the Gain, Discrepancy, and Top-$k$ rule strategies, we evaluate three approaches to relax the rules between restarts: \textit{Monotonic}, where the threshold is increased by a fixed amount (set to~$1$ in our experiments); \textit{Exponential}, where the threshold is multiplied by a constant factor ($2$ in our experiments). We also use \textit{Luby}, where the increment follows the Luby sequence from~\cite{lorenz2021restart}.
Across both depths, all complete anytime strategies outperform DL8.5, highlighting the benefits of rule-based restarting. The best overall results are achieved by CA-Discrepancy (Luby and Monotonic) and CA-Top-$k^*$, especially under longer time budgets. At depth~$6$, CA-Discrepancy with monotonic relaxation achieves the lowest average primal integral of $7.4$ at 300s, while CA-Top-$k^*$ reaches $7.7$. 

Under short timeouts ($15–30$s), Blossom produces high quality early solutions, often outperforming CADL8.5 variants. However, CADL8.5 quickly catches up and surpasses Blossom as runtime increases. This trend becomes more pronounced at depth~7 (Table~\ref{tab:d7_p}), where CA-Discrepancy with Monotonic and Luby relaxation achieves average primal integral values of $14.0$, better than Blossom's $17.2$ at $300$s. These improvements are largely due to the increased diversification in its search strategy, exploring more parts of the search space, whereas Blossom tends to remain focused on optimizing the deeper layers of a specific tree before moving elsewhere.

Among all rules, \emph{Discrepancy} consistently outperforms the others. CA-Top-$k$ and its variant CA-Top-$k^*$ also perform well, particularly at large timeouts. CA-Gain and CA-Purity lag behind: the Gain rule often selects larger subtrees, leading to longer subsearches; Purity may require several ineffective relaxations before contributing to meaningful diversification.

Greedy baselines such as C4.5, Top-$3$, and Top-$5$ deliver quick but static solutions. Among them, Top-$5$ performs best under tight time budgets ($15–30$s), briefly outperforming DL8.5. However, none of the greedy methods improve their solutions over time.

\begin{figure}[ht]
    \centering
    \includegraphics[width=0.9\linewidth]{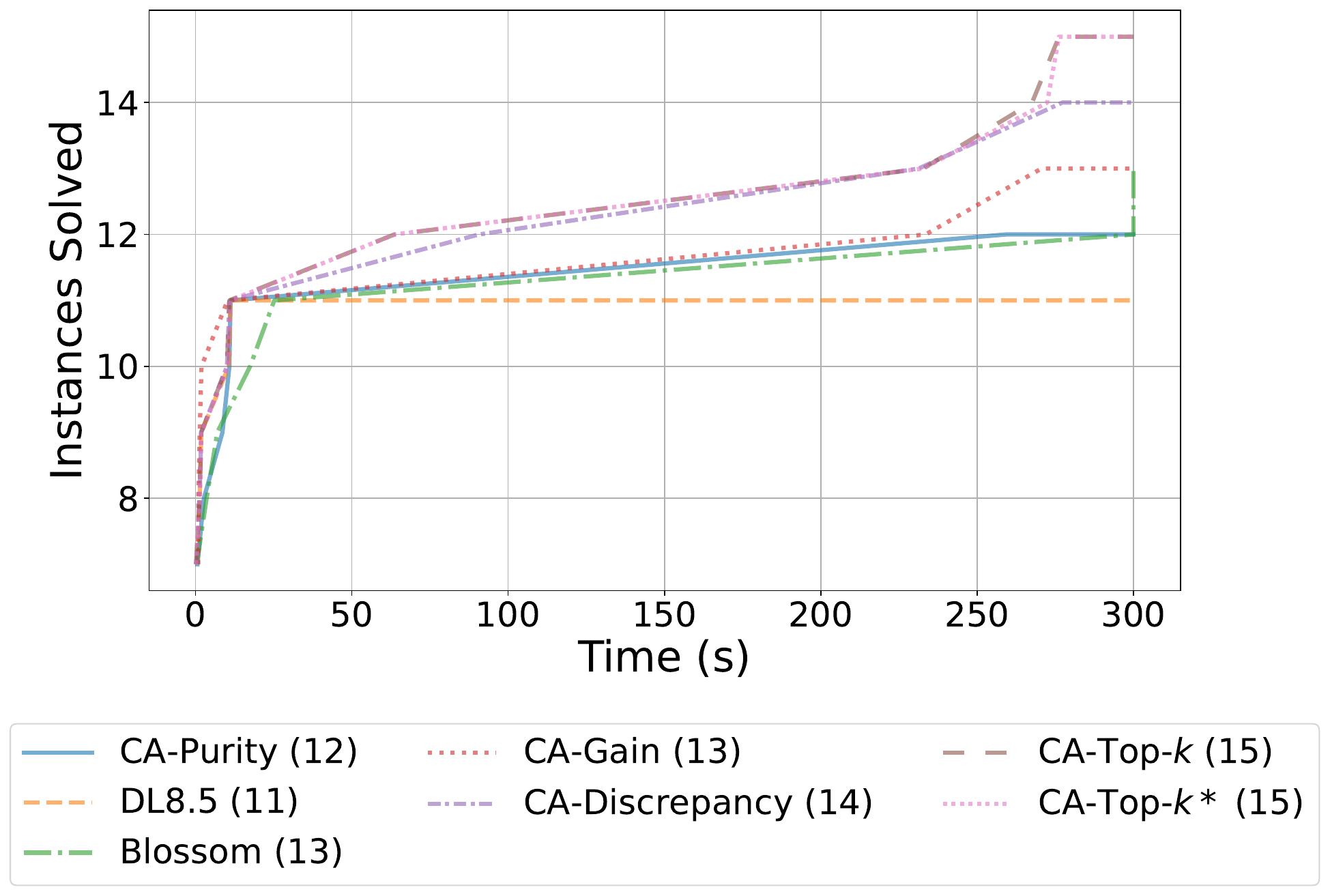}
    \caption{Cumulative number of instances solved as a function of time with the total number of instances solved by each approach within $300$s}
    \label{fig:cumplot}
\end{figure}
Figure~\ref{fig:cumplot} illustrates the cumulative termination count of each algorithm to find and prove optimality within a time budget of $300$s and a depth constraint of $5$.
Overall, the CADL8.5 variants demonstrate superior solving power compared to DL8.5 and Blossom. Notably, CA-Top-$k^*$ solves the most instances ($15$ out of $25$) and does so more quickly than the other methods across most of the timeline. CA-Discrepancy and CA-Top-$k$ also perform strongly, solving $14$ and $15$ instances respectively, and surpassing other approaches beyond the 50-second mark. Blossom shows a steep initial rise, indicating strong early performance, but plateaus sooner than the CADL8.5 variants. CA-Gain achieves a similar final count as Blossom ($13$ instances) but shows slower progress in the early phase. DL8.5 and CA-Purity underperform both in terms of speed and total solved instances, solving only $11$ and $12$ datasets respectively. This shows that CADL8.5 variants especially Top-$k$(*) and Discrepancy does not compromise the ability to reach optimal solutions.

\section{Conclusion}
In this paper, we introduced CADL8.5, a complete anytime framework for decision tree learning that generalizes DL8.5, LDS-DL8.5, and Top-$k$. It combines DL8.5's efficient branch-and-bound pruning and trie-based caching with a restart-based search that progressively relaxes pruning criteria, guided by rule based strategies such as node purity, Information Gain gap, Discrepancy and Top-$k$. Our experiments show that CADL8.5 variants, especially CA-Top-$k^*$ and CA-Discrepancy, deliver strong anytime performance without sacrificing the ability to reach optimal solutions. They solve more instances to optimality than DL8.5 and Blossom and perform at least on par with greedy approaches, while improving solution quality over time. Future work includes exploring combined rule strategies such as Gain and Discrepancy.

\bibliography{aaai2026}

\end{document}